\documentclass{article}
\usepackage[preprint,nonatbib]{nips_2018}
\pdfoutput=1
\usepackage[utf8]{inputenc} 
\usepackage[T1]{fontenc}    
\usepackage{hyperref}       
\usepackage{url}            
\usepackage{booktabs}       
\usepackage{amsfonts}       
\usepackage{amssymb}
\usepackage{amsmath}
\usepackage{nicefrac}       
\usepackage{microtype}      
\usepackage{graphicx}
\usepackage{bmpsize}
\usepackage{bm}
\usepackage{booktabs} 
\usepackage{wrapfig}
\usepackage{caption}
\usepackage{xcolor}
\usepackage{todonotes}

\begin{document}
\title{The Actor Search Tree Critic (ASTC) for Off-Policy POMDP Learning in Medical Decision Making}
\author{Luchen Li\\
Imperial College London\\
l.li17@imperial.ac.uk
\And Matthieu Komorowski\\
Imperial College London\\
matthieu.komorowski@gmail.com
\And A. Aldo Faisal\\
Imperial College London\\
a.faisal@imperial.ac.uk}

\maketitle
\begin{abstract}
Off-policy reinforcement learning enables near-optimal policy from suboptimal experience, thereby provisions opportunity for artificial intelligence applications in healthcare. Previous works have mainly framed patient-clinician interactions as Markov decision processes, while true physiological states are not necessarily fully observable from clinical data. We capture this situation with partially observable Markov decision process, in which an agent optimises its actions in a belief represented as a distribution of patient states inferred from individual history trajectories. A Gaussian mixture model is fitted for the observed data. Moreover, we take into account the fact that nuance in pharmaceutical dosage could presumably result in significantly different effect by modelling a continuous policy through a Gaussian approximator directly in the policy space, i.e. the actor. To address the challenge of infinite number of possible belief states which renders exact value iteration intractable, we evaluate and plan for only every encountered belief, through heuristic search tree by tightly maintaining lower and upper bounds of the true value of belief. We further resort to function approximations to update value bounds estimation, i.e. the critic, so that the tree search can be improved through more compact bounds at the fringe nodes that will be back-propagated to the root. Both actor and critic parameters are learned via gradient-based approaches. Our proposed policy trained from real intensive care unit data is capable of dictating dosing on vasopressors and intravenous fluids for sepsis patients that lead to the best patient outcomes.

\end{abstract}

%
%



\section{Introduction}
Many recent examples \cite{gulshan2016development, esteva2017dermatologist, litjens2017survey} have demonstrated that machine learning can deliver above-human performance in classification-based diagnostics. However, key to medicine is diagnosis paired with treatment, i.e. the sequential decisions that have to be made by clinician for patient treatment. 
Automatic treatment optimisation based on reinforcement learning has been explored in simulated patients for HIV therapy \cite{10Ernst2006CDB} using fitted-Q iteration, dynamic insulin dosage in diabetes using model-based reinforcement learning \cite{11Bothe2013TheUO}, and anaesthesia depth control using actor-critic \cite{Lowery2013}.


We focus on principled closed-loop approaches to learn a near-optimal treatment policy from vast electronic healthcare records (EHRs). Previous works mainly modelled this problem as a Markov decision process (MDP) and learned a policy of actions $\pi(a|s)=P(a|s)$ based on estimated values $v(s)$ of possible actions $a$ in each state $s$.
\cite{8Shortreed2011ISCD} investigated fitted Q-iteration with linear function approximation to optimise treatment of schizophrenia. 
\cite{12Prasad2017MVI} compared different supervised learning methods to approximate action values, including neural networks, and successfully predicted the weaning of mechanical ventilation and sedation dosages. In addition to crafting a reward function that reflects domain knowledge, another avenue explored inverse reinforcement learning to recover one from expert behaviours. For example, \cite{19Asoh2013AAI} proposed hemoglobin-A1c dosages for diabetes patients by implementing Markov chain Monte Carlo sampling to infer posterior distribution over rewards given observed states and actions. Further, multiple objective optimal treatments were explored \cite{25Lizotte2016MOMD} with non-deterministic fitted-Q to provide guidelines on antipsychotic drug treatments for schizophrenia patients. \cite{24Nemati2016OptimalMD} inferred hidden states via discriminative hidden Markov model, and investigated a Deep-fitted Q variant to learn heparin dosages to manage thrombosis.

However, the true  physiological state of the patient is not necessarily fully observable by clinical measurement techniques. Practically, this observability is further restricted by the actual subset of measures taken in hospital, restricting both the nature and frequency of data recorded 
and omitting information readily visible to the clinician 
. Those latencies are especially salient in intensive care units (ICUs). Strictly speaking, the decision making process acting on the patient state therefore should be mathematically formulated as a partially observable MDP (POMDP). Consequently, the true patient state is latent and thus the patient's \emph{physiological belief state} $b$ can only be represented as a probability distribution of the states given the history of observations and actions. Exhaustive and exact POMDP solutions are computationally intractable because belief states constitute a continuous hyperplane that contains infinite possibilities.
We alleviate this obstruct by evaluating the sequentially encountered belief states through learning both upper and lower bounds of their true values and exploring reachable sequences using heuristic search trees to obtain locally optimal value estimates $\hat{v}_{\tau}(b)$. The learning of value bounds in the tree search is conferred via keeping back-propagating that of newly expanded nodes to the root, which are usually computed offline and remain static. We model the lower and upper value bounds as function approximators so that the tree search can be improved through more accurate value bounds at the fringe nodes as we process the incoming data. Moreover, to realise continuous action spaces (e.g. milliliter of medicine dripped per hour) we explicitly use a continuous policy implemented through function approximation. We embed these features in an actor-critic framework to update the parameters of the policy (the actor) and the value bounds (the critic) via gradient-based methods.

\section{Preliminaries}
\paragraph{POMDP Model}
A POMDP framework can be represented by the tuple $\left\{\mathcal{S,A,O,T}, \Omega, \mathcal{R}, \gamma, b_{0} \right\}$ \cite{29Pineau2003PBVI}, where $ \mathcal{S} $ is the state space, $ \mathcal{A} $ the action space, $ \mathcal{O} $ the observation space, $ \mathcal{T} $ the stochastic state transition function: $ \mathcal{T}(s', s, a) = P(s_{t+1}=s'|s_{t}=s, a_{t}=a)$, $ \Omega $ the stochastic observation function: $ \Omega(o, s, a) = P(o_{t+1}=o|s_{t+1}=s, a_{t}=a)$, $ \mathcal{R} $ the immediate reward function: $ r=\mathcal{R}(s, a) $, $ \gamma \in [0, 1) $ the discount factor indicating the weighing of present value of future rewards, and $b_{0}$ the agent's initial knowledge before receiving any information.

\begin{figure}[!htp]
 \centerline{
		\begin{tabular}{cc}
			\includegraphics[scale=1,width=0.48\textwidth]{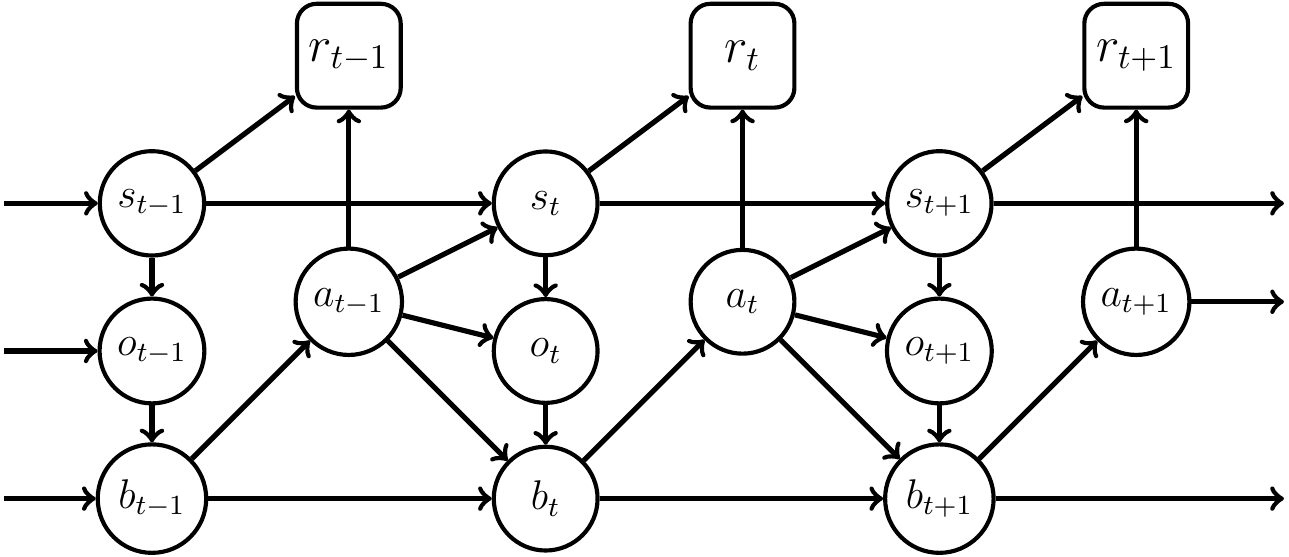} 
			&
			\includegraphics[width=0.48\textwidth]{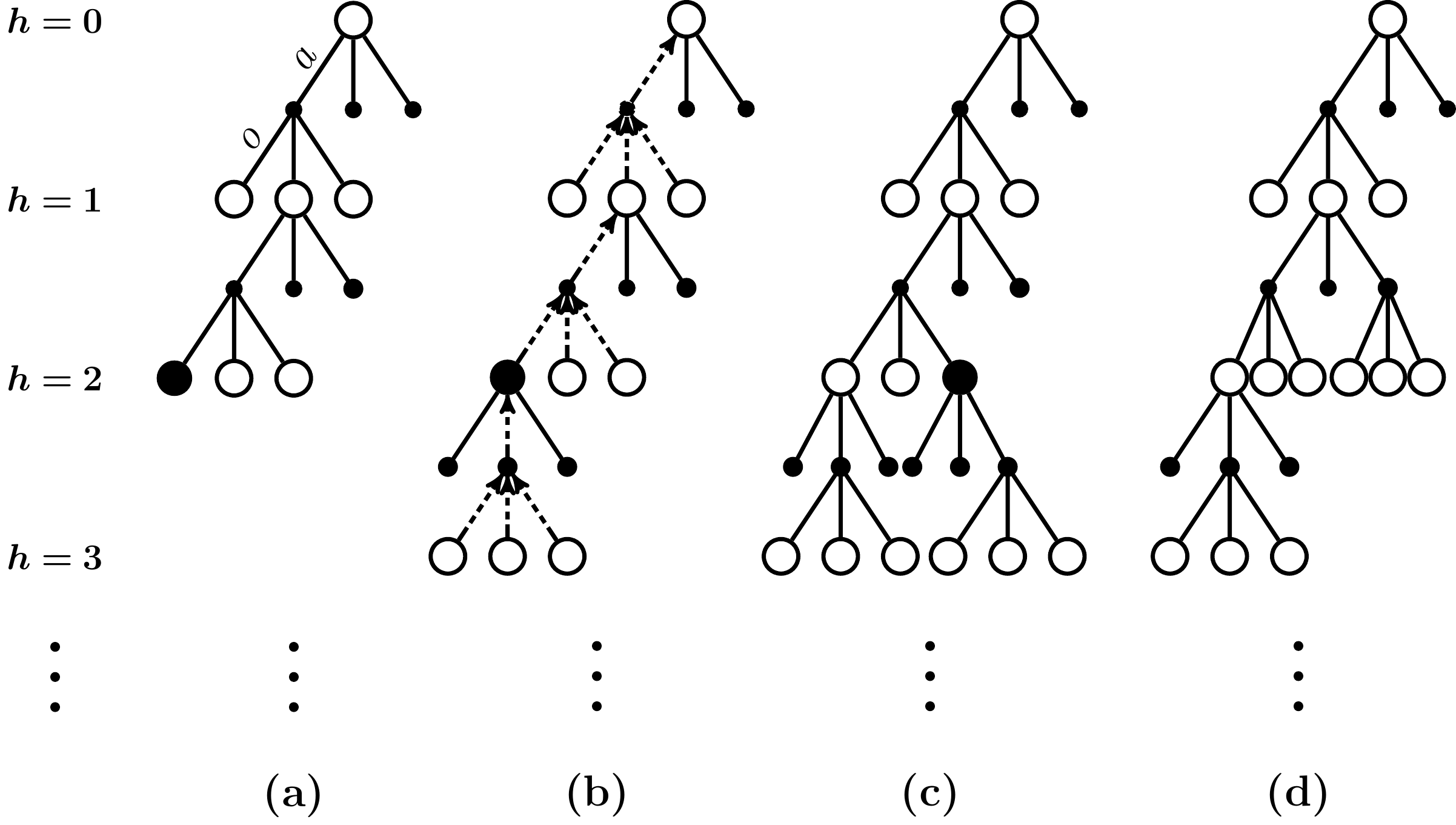}\\
		\end{tabular}
	}
  \caption{\textbf{(left)} Graphical model of POMDP. \textbf{(right)} Tree search. Each circle node represents a belief state, each dotted node an action-belief-state pair. (a) select the best fringe node to expand; (b) expand the selected node by choosing the action with the maximal upper value bound and considering all possible observations, and back-propagate value bounds through all its ancestors to the root; (c) no revision is required for previous choices on actions, expand the next best fringe node; (d) a previous action choice (example is showing in horizon 1) is no longer optimal after the latest expansion, the new optimal action is selected.}
\label{fig:gm_pomdp_tree}
\end{figure}

The agent's belief state is represented by the probability distribution over states given historical observations and actions,  $ b_{t}(s) = P(s~|~b_{0}, o_{1}, a_{1}, ..., a_{t-1}, o_{t}) $. Executing $ a $ in belief $ b $ and receiving observation $ o $, the new belief is updated through:
\begin{equation}\label{eq:belief_trans}
b'(s') = \tau(b, a, o) = \eta\Omega(o, s', a)\sum_{s\in\mathcal{S}}{\mathcal{T}(s', s, a)b(s)}
\end{equation}
where $ \eta = \frac{1}{P(o|b, a)} $ is a normalisation constant, and
\begin{equation}
P(o|b, a) = \sum_{s'\in\mathcal{S}}{\Omega(o, s', a)\sum_{s\in\mathcal{S}}{\mathcal{T}(s', s, a)b(s)}}
\end{equation}
The optimal policy $ P(a|b)=\pi (a|b) $ specifies the best action to select in a belief, its value function being updated via the fixed point of Bellman equation \cite{Bellman1957}:
\begin{equation}\label{eq:opt_value}
v^*(b) = \max_{a\in\mathcal{A}}  \left[ \mathcal{R}_{B}(b, a)+\gamma\sum_{o\in\mathcal{O}} P(o|b, a)v^*\left(\tau \left(b, a, o\right)\right) \right]
\end{equation}
$ \mathcal{R}_{B}(b, a) = \sum_{s\in\mathcal{S}}\mathcal{R}(s, a)b(s) $ is the probability-weighted immediate reward. Graphical model of a fraction of POMDP framework is shown in Fig.~\ref{fig:gm_pomdp_tree} (left).

\paragraph{Continuous Control}
Policy gradient method can go beyond the limit of a finite action space and achieve continuous control. Instead of choosing actions based on action-value estimates, a policy is directly optimised in the policy space. Our policy $ \pi(a|s, \bm{u}) = \mathcal{F}_{actor}(\bm{x}(s), \bm{u}) $ is a function of state feature vector $\bm{x}(s)$ parameterised by weight vector $ \bm{u} $. The objective function measuring the performance of policy $ \pi $ is defined as the value, or expected total rewards from future, of the start state:
\begin{equation}
J(\pi)=E_{\pi}\left[G_{0}|s_{0}\right]=E_{\pi}\left[\sum_{t=0}^{T-1}\gamma^{t}r_{t}~|~s_{0}\right]
\end{equation}
where $ G_{0} $ is the return $ G_{t} = \sum_{k=0}^{T-t-1}\gamma^{k}r_{t+k} $ at $ t=0 $. According to policy gradient theorem \cite{62Sutton1999PGM}, the gradient of $ J $ w.r.t. $ \bm{u} $ is:
\begin{equation} \label{eq:grad_theorem}
\nabla_{\bm{u}}J(\pi) = \footnote{$\nabla_{\bm{u}}J(\pi) = \sum_{s\in\mathcal{S}}P_{\pi}^{stat}(s)\sum_{a\in\mathcal{A}} q_{\pi}(s_{t}, a)\nabla_{\bm{u}}\pi(a|s_{t}, \bm{u}_{t})
= E_{\pi}\left[\sum_{a\in\mathcal{A}} q_{\pi}(s_{t}, a)\nabla_{\bm{u}}\pi(a|s_{t}, \bm{u}_{t})\right]
= E_{\pi}\left[\sum_{a\in\mathcal{A}} \pi(a|s_{t}, \bm{u}_{t})q_{\pi}(s_{t}, a)\frac{\nabla_{\bm{u}}\pi(a|s_{t}, \bm{u}_{t})}{\pi(a|s_{t}, \bm{u}_{t})}\right]
= E_{\pi}\left[q_{\pi}(s_{t}, a_{t})\frac{\nabla_{\bm{u}}\pi(a_{t}|s_{t}, \bm{u}_{t})}{\pi(a_{t}|s_{t}, \bm{u}_{t})}\right] = E_{\pi}\left[G_{t}\frac{\nabla_{\bm{u}}\pi(a_{t}|s_{t}, \bm{u}_{t})}{\pi(a_{t}|s_{t}, \bm{u}_{t})}\right]$.} ~~ \sum_{s\in\mathcal{S}}P_{\pi}^{stat}(s)\sum_{a\in\mathcal{A}} q_{\pi}(s_{t}, a)\nabla_{\bm{u}}\pi(a|s_{t}, \bm{u}_{t}) = E_{\pi}\left[G_{t}\frac{\nabla_{\bm{u}}\pi(a_{t}|s_{t}, \bm{u}_{t})}{\pi(a_{t}|s_{t}, \bm{u}_{t})}\right]
\end{equation}
$P_{\pi}^{stat}(s)$ denotes the stationary distribution of states under policy $\pi$ (i.e. the chance a state will be visited within an episode). $q_{\pi}(s_{t}, a_{t}) = E_{\pi}\left[G_{t}|s_{t}, a_{t}\right]$ is the value of $a_{t}$ in state $s_{t}$. $\bm{u} $ can subsequently be updated through gradient ascent in the direction (with step size $ \alpha $) that maximally increases $ J $:
\begin{equation}\label{eq:actor_update_raw}
\bm{u}_{t+1} = \bm{u}_{t}+\alpha G_{t}\frac{\nabla_{\bm{u}}\pi(a_{t}|s_{t}, \bm{u}_{t})}{\pi(a_{t}|s_{t}, \bm{u}_{t})}
\end{equation}

\section{Methodology}
\begin{wrapfigure}{l}{0.62\textwidth}
  \begin{center}
    \includegraphics[width=0.6\textwidth]{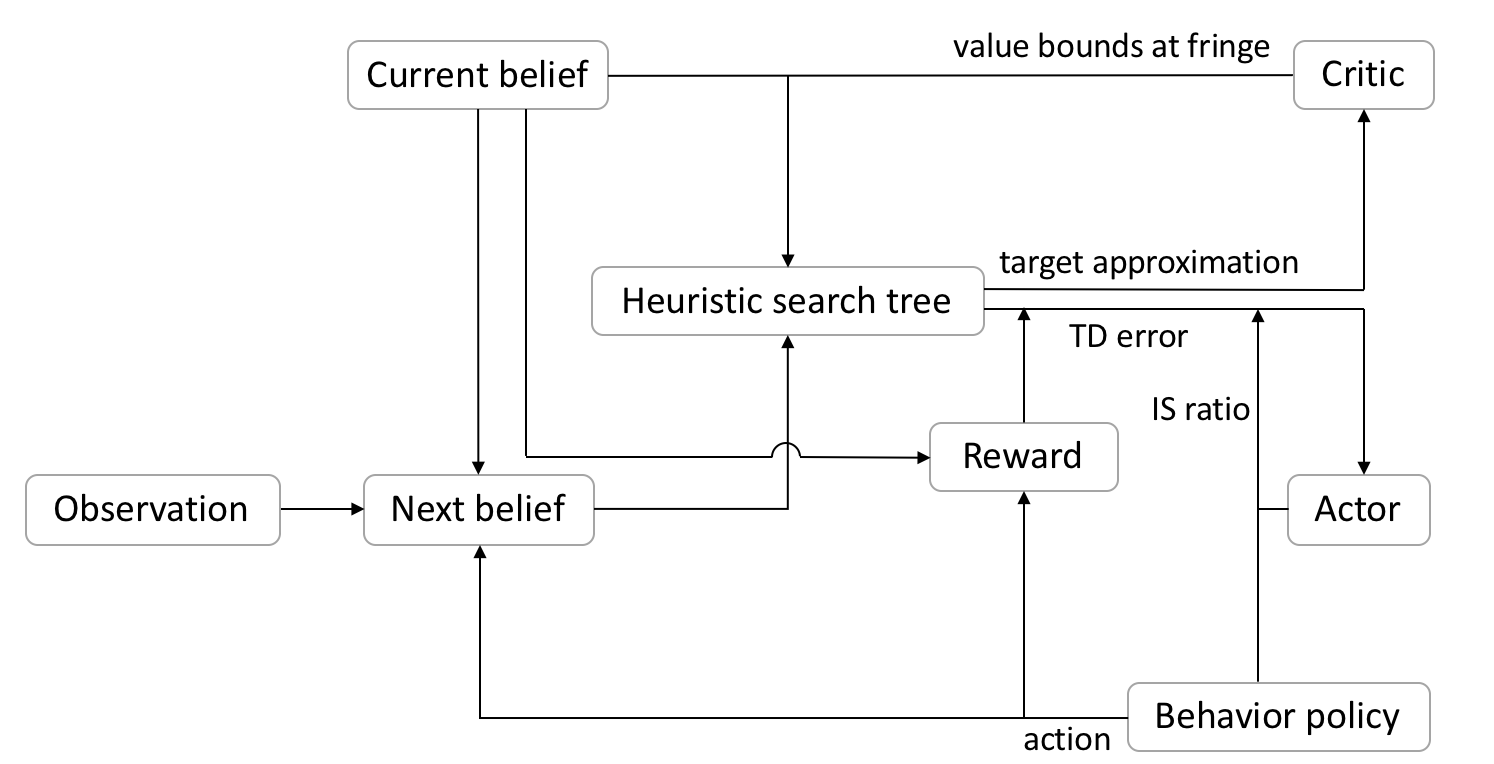}
  \end{center}
  \caption{ASTC algorithm flowchart.}
  \label{fig:alg}
\end{wrapfigure}

In this section, we introduce how our algorithm combines tree search with policy gradient/function approximations to realise both continuous action space and efficient online planning for POMDP. We will derive the components of our algorithm as we go along this section and provide an overview flowchart in Fig.~\ref{fig:alg}.
\paragraph{Heuristic Search in POMDPs}
The complexity of solving POMDPs is mainly due to the \textit{curse of dimensionality}: a belief state in an $ |\mathcal{S}| $ space is an $ (|\mathcal{S}|-1 )$-dimensional continuous simplex, with all its elements sum to one, and \textit{curse of history}, as it acknowledges previous observations and actions, the combination of which grows exponentially with the planning horizon \cite{29Pineau2003PBVI}. Exact value iteration for Eq.~\ref{eq:opt_value} in conventional reinforcement learning is therfore computationally intractable.

The optimal value function $ v^*(b) $ of a finite-horizon POMDP is piecewise linear and convex in the belief state \cite{30Sondik1978TOCP}, represented by a set of $ |\mathcal{S}| $-dimensional convex hyperplanes, whose total amount grows exponentially. Most exhaustive algorithms for POMDPs are dedicated to learning either lower bound \cite{29Pineau2003PBVI, 39Spaan2004APPOMDP} or upper bound \cite{33Littman1995LPP, 35Cassandra1994AOI, 40Hauskrecht2000VAP} of $ v^*(b) $ by maintaining a subset of the aforementioned hyperplanes. Tree-search based solutions usually prunes away less likely observations or actions \cite{43Paquet06hybridpomdp, 44McAllester1999APF, 45Kearns2002SSA}, or expanding fringe nodes according to a predefined heuristic \cite{42Smith2004HSV, 51Washington97bi-pomdp:bounded, 52Ross2007AAO}.

Our methodology is compatible with any tree-search based POMDP solution. Here we implement on a heuristic tree search introduced in \cite{52Ross2007AAO} to focus computations on every encountered belief (i.e. plan at decision time) and explore only reachable 
sequences. Specifically, a search tree rooted at the current belief $b_{curr}$ is built, whose value 
estimate is confined by its lower bound $\hat{v}^{L}(b_{curr})$ and upper bound $\hat{v}^{U}(b_{curr})$ 
that after each step of look-ahead become tighter to the true optimal value $ v^*(b_{curr}) $. Here we 
use subscript on the belief state to denote its temporal position within the episode in environmental 
experience, and superscript the horizon explored for it in the tree search (analogous for observation 
and action). At each update during exploration, only the fringe node that leads to the maximum error on 
the root $b^{0}_{curr}$ is expanded:

\begin{equation}
b^*_{curr} = \mathop{\mathrm{arg\,max}}_{b^{h}_{curr}, ~h\in\left\{0, ..., H\right\}} \left[\gamma^{h}\left[\hat{v}^{U}(b^{h}_{curr})-\hat{v}^{L}(b^{h}_{curr})\right]\prod_{i=0}^{h-1}P(o^{i}_{curr}|b^{i}_{curr}, a^{i}_{curr})\pi_{\tau}(a^{i}_{curr}|b^{i}_{curr})\right]
\end{equation}

where $H$ is the maximum horizon explored so far, $\hat{v}^{U}(b^{h}_{curr})-\hat{v}^{L}(b^{h}_{curr}) $ the error on the fringe node, $ \prod_{i=0}^{h-1}P(o^{i}_{curr}|b^{i}_{curr}, a^{i}_{curr})\pi_{\tau}(a^{i}_{curr}|b^{i}_{curr}) $ the probability of reaching it from the root, and $ \gamma^{h} $ the time discount (Fig.~\ref{fig:gm_pomdp_tree} (right) (a)). Once a node is expanded, the estimates of value bounds of all its ancestors are updated in a bottom-up fashion analogous to equation Eq.~\ref{eq:opt_value}, substituting $v^*$ with $\hat{v}^{L}$ or $\hat{v}^{U}$ as appropriate, to the root (Fig.~\ref{fig:gm_pomdp_tree} (right) (b)), and previous choices on actions in expanded belief nodes along the path are revised (Fig.~\ref{fig:gm_pomdp_tree} (right) (c-d)) to ensure that the optimal action $a^{i*}_{curr}$ in $b^{i}_{curr}$ for $\forall ~h\in\left\{0, ..., H\right\}$ is always explored based on current estimates.
\begin{equation}\label{eq:tree_policy}
a^{i*}_{curr} = \mathop{\mathrm{arg\,max}}_{a\in\mathcal{A}} \left[\mathcal{R}_{B}(b^{i}_{curr}, a)+\gamma\sum_{o\in\mathcal{O}} P(o|b^{i}_{curr}, a)\hat{v}^{U}\left(\tau (b^{i}_{curr}, a, o)\right)\right]
\end{equation}

Eq.~\ref{eq:tree_policy} is the deterministic tree policy $\pi_{\tau}(a|b)$ that guides exploration within the tree. Each expansion leads to more compact value bounds at the root. Tree exploration is terminated when the interval between the value bounds estimates at the root belief changes trivially or a time limit is reached.

\paragraph{Gaussian States}
Observed patient information is modelled as a Gaussian mixture, each observation being generated from one of a finite set of Gaussian distributions that represent genuine physiological states. The total number of latent states is decided by Bayesian information criterion \cite{Schwarz1978BIC} through cross-validation using the development set. The terminal state is observable and corresponds either patient discharge or death. Eq.~\ref{eq:belief_trans} can be further expressed as:
\begin{equation}
b'(s') =  \eta\frac{P^{a}(s'|o)P^{a}(o)}{P^{a}(s')}\sum_{s\in\mathcal{S}}{\mathcal{T}^{a}(s', s)b(s)} = \eta'\frac{P^{a}(s'|o)}{P^{a}(s')}\sum_{s\in\mathcal{S}}{\mathcal{T}^{a}(s', s)b(s)}
\end{equation}
The superscript $ a $ denotes a subset divided from the data according to the action taken during this 
transition \footnote{In our implementation, $ P^{a}(s'|o) $ for $ \forall a\in\mathcal{A} $ are 
globally computed as $ P(s'|o) \forall o\in\mathcal{O} $, regardless of the action leading to 
it, because the impact of the observation on the distribution of states is significantly more substantial than the action administered.}. The division into subsets enables parallel computations. $ P^{a} (s'|o) $ is the posterior distribution of $ s' $ when observing $ o $ speculated from the trained Gaussian mixture model. 

The transition function $ \mathcal{T}^{a}(s', s) $ is learned by maximum a posteriori 
\cite{Murphy2012MLA} to allow possibilities for transitions that did not occur in the development dataset.

Prior knowledge on transitions is modelled by Griffiths-Engen-McCloskey (GEM) distributions 
\cite{63buntine2012ABVOP} according to relative Euclidean distances between state centroids, whose 
elements, if sorted, decrease exponentially and sum to one, to reflect higher probabilities of 
transiting into similar states.  Specifically, a GEM distribution is defined by a discount parameter $c_{1}$ and a concentration  parameter $c_{2}$, and can be explained by a stick-breaking construction: break a stick for the $k$-th time  into two parts, whose length proportions conform to a Beta distribution:
\begin{equation}
V_{k}\sim \mathrm{Beta}(1-c_{1}, c_{2}+kc_{1}), ~~~~
0\leq c_{1}<1, ~c_{2}>-c_{1}
\end{equation}
Then the length proportions of the off-broken parts in the whole stick are:
\begin{equation}\label{eq:stick_break}
p_{k} = \begin{cases}
V_{1}, ~~~~ k=1\\
(1-V_{1})(1-V_{2})...(1-V_{k-1})V_{k}, ~~~~ k=2,3,...
\end{cases}
\end{equation}
The probability vector $\bm{p}$ consisting of elements calculated from Eq.~\ref{eq:stick_break} constitutes a GEM distribution.

\paragraph{Actor-Critic}
Including to Eq.~\ref{eq:grad_theorem} a baseline term $ B(s_{t}) $ as a comparison with $ G_{t} $ 
significantly reduces variance in gradient estimates without changing equality \footnote{Because $ 
\sum_{a\in\mathcal{A}}B(s)\nabla_{\bm{u}}\pi(a|s, 
\bm{u})=B(s)\nabla_{\bm{u}}\sum_{a\in\mathcal{A}}\pi(a|s, \bm{u})=B(s)\nabla_{\bm{u}}1=0 $, $ \forall 
s\in\mathcal{S} $.}. This baseline is required to discern states, a natural candidate would be the state 
value or its parametric approximation $ \hat{v}(s, \bm{w}) = \mathcal{F}_{critic}(\bm{x}(s), \bm{w}) 
$. Then Eq.~\ref{eq:actor_update_raw} becomes:
\begin{equation}\label{eq:actor2}
\bm{u}_{t+1} = \bm{u}_{t}+\alpha \left[G_{t}-\hat{v}(s_{t}, \bm{w}_{t})\right]\frac{\nabla_{\bm{u}}\pi(a_{t}|s_{t}, \bm{u}_{t})}{\pi(a_{t}|s_{t}, \bm{u}_{t})}
\end{equation}
The parametric policy $ \pi(a|s, \bm{u}) $ is called the \textit{actor}, and the parametric value function $ \hat{v}(s, \bm{w}) $ the \textit{critic}.

The complete empirical return $ G_{t} $ is only available at the end of each episode, and therefore at 
each update we need to look forward to future rewards to decide current theoretical estimate of $ G_{t} 
$. A close approximation of $ G_{t} $ that is available at each decision moment and thus enables more 
data-efficient backward-view learning is $\lambda$-return: $ G_{t}^{\lambda} = (1-
\lambda)\sum_{n=1}^{\infty}\lambda^{n-1}G_{t:t+n} $, with $ \lambda $ specifying the relative decaying 
rate among returns available after various steps $ G_{t:t+n} $. Since $ G_{t}^{\lambda}-\hat{v}(s_{t}, 
\bm{w}_{t}) \approx r_{t}+\gamma \hat{v}(s_{t+1}, \bm{w}_{t})-\hat{v}(s_{t}, \bm{w}_{t}) $ (denoted as $ 
\delta_{t} $), substituting $ G_{t} $ with $ G_{t}^{\lambda} $ in Eq.~\ref{eq:actor2} yields:
 \begin{equation}\label{eq:actor3}
 \bm{u}_{t+1} = \bm{u}_{t}+\alpha \delta_{t}\bm{e}_{t}^{\bm{u}}
 \end{equation}
\begin{equation}\label{eq:eligi1}
\bm{e}_{t}^{\bm{u}} = \gamma\lambda \bm{e}_{t-1}^{\bm{u}}+\frac{\nabla_{\bm{u}}\pi(a_{t}|s_{t}, \bm{u}_{t})}{\pi(a_{t}|s_{t}, \bm{u}_{t})}
\end{equation}
$ \bm{e}^{\bm{u}} $ is the eligibility trace \cite{Klopf1986} for $ \bm{u} $, and is initialised to $ \bm{0} $ for every episode. Analogous update rules apply to the critic.

\paragraph{Actor Search Tree Critic}
The history-dependent probabilistic \emph{belief state} reflects the information the agent would need to 
know about the current time step to optimise its decision. This belief state is used as the state feature vector for the actor-critic, i.e. $\bm{x} (s) = b$. This is based on the notion that state mechanism is supposed to allow weight parameter to update towards similar directions by similar samples, while similar situations have similar distributions of states, with each component in the distribution implying the responsibility for updating corresponding component in the weight.


Our critic parameterises the lower and upper value bounds in the tree search, instead of parameterising 
the value function as whole (as done in conventional actor-critic methods). Note that the value bounds 
at the fringe nodes are here updated as we parse the data for off-policy reinforcement 
learning. In previous works these bounds would have been computed offline and not improved during online 
planning. We use linear representations for the value bounds: 
\begin{equation}\label{eq:linear_v_bounds}
\hat{v}^{L}(b, \bm{w}^{L}) = \bm{w}^{L~T}b, ~~~~ \hat{v}^{U}(b, \bm{w}^{U}) = \bm{w}^{U~T}b
\end{equation}

At each step $t$, a local (as opposed to a value function optimal to all belief states) optimal value is estimated for the current belief $ b_{t} $ through heuristic tree search with fringe nodes values approximated by $ \bm{w}^{L} $ and $ \bm{w}^{U} $, denoted as $ \hat{v}_{\tau}(b_{t}, \bm{w}^{L}_{t}, \bm{w}^{U}_{t}) $. The critic parameters are updated through stochastic gradient descent (SGD) to adjust in the direction that most reduces the error on each training example by minimising the mean square error between the current approximation and its target $ \hat{v}_{\tau}(b_{t}, \bm{w}^{L}_{t}, \bm{w}^{U}_{t}) $:
\begin{equation}
\bm{w}^{L}_{t+1} = \footnote{$\bm{w}^{L}_{t+1} =\\ \bm{w}^{L}_{t}-\frac{1}{2}\beta^{L}\nabla_{\bm{w}}\left[\hat{v}_{\tau}(b_{t}, \bm{w}^{L}_{t}, \bm{w}^{U}_{t})-\hat{v}^{L}(b_{t}, \bm{w}^{L})\right]^{2} = \bm{w}^{L}_{t}+\beta^{L}\left[\hat{v}_{\tau}(b_{t}, \bm{w}^{L}_{t}, \bm{w}^{U}_{t})-\hat{v}^{L}(b_{t}, \bm{w}^{L})\right]\nabla_{\bm{w}}\hat{v}^{L}(b_{t}, \bm{w}^{L}) = \bm{w}^{L}_{t}+\beta^{L}\left[\hat{v}_{\tau}(b_{t}, \bm{w}^{L}_{t}, \bm{w}^{U}_{t})-\hat{v}^{L}(b_{t}, \bm{w}^{L})\right]b_{t}$.} ~~ \bm{w}^{L}_{t}+\beta^{L}\left[\hat{v}_{\tau}(b_{t}, \bm{w}^{L}_{t}, \bm{w}^{U}_{t})-\hat{v}^{L}(b_{t}, \bm{w}^{L})\right]b_{t}
\end{equation}
$\beta^{L}$ is the step size for updating $\bm{w}^{L}$. Similar for $ \bm{w}^{U}_{t} $. As the weight vector also has an impact on the target $ \hat{v}_{\tau}(b_{t}, \bm{w}^{L}_{t}, \bm{w}^{U}_{t}) $, which is ignored during SGD update, the update is by definition \emph{semi-gradient}, which usually learns faster than full gradient methods and with linear approximators (Eq.~\ref{eq:linear_v_bounds}), is guaranteed to converge (near) to a local optimum
under standard stochastic approximation conditions \cite{9Sutton98a}:
\begin{equation}
\sum_{t=1}^{\infty}\beta_{t}=\infty,~~~~
\sum_{t=1}^{\infty}\beta_{t}^{2}<\infty
\end{equation}
To ensure convergence, we set the step sizes at time step $t$ according to:
\begin{equation}
\beta^{L}_{t} = \frac{0.1}{E\left[\parallel b_{t}\parallel _{\bm{w}^{L}_{t}}^{2}\right]}, ~~~~
\beta^{U}_{t} = \frac{0.1}{E\left[\parallel b_{t}\parallel _{\bm{w}^{U}_{t}}^{2}\right]}
\end{equation}


To realise continuous action space, the actor is modelled as a Gaussian distribution, with a mean vector approximated as a linear function (for simplicity of gradient computation) of weights and belief state:
\begin{equation}
\pi(a|b, \bm{u}) = \mathcal{N}(\bm{u}^{T}b, ~\sigma^{2})
\end{equation}
$\sigma$ is a hyperparameter for standard deviation. In this circumstance the gradient in Eq.~\ref{eq:eligi1} is calculated as:
\begin{equation}
\frac{\nabla_{\bm{u}}\pi(a|b, \bm{u})}{\pi(a|b, \bm{u})} = \frac{1}{\sigma^{2}}(a-\bm{u}^{T}b)b
\end{equation}
The moment-wise error, or temporal difference (TD) error that motivates update to actor in Eq.~\ref{eq:actor3} is:
\begin{equation}
\delta_{t} = r_{t}+\gamma\hat{v}_{\tau(b_{t+1}, \bm{w}^{L}_{t}, \bm{w}^{U}_{t})}-\hat{v}_{\tau(b_{t}, \bm{w}^{L}_{t}, \bm{w}^{U}_{t})}
\end{equation}

In off-policy reinforcement learning, as we use retrospective data generated by a behaviour policy $ \pi_{b} $ (being clinicians' actual treatment decisions) to optimise a target policy $ \pi $ (being our actor), the actor-critic approach is tuned via importance sampling on the eligibility trace \cite{66Degris2012OffPolicyA}:
\begin{equation}
 \bm{e}_{t}^{\bm{u}} = \rho_{t}\left[ \gamma\lambda \bm{e}_{t-1}^{\bm{u}}+\frac{\nabla_{\bm{u}}\pi(a_{t}|b_{t}, \bm{u}_{t})}{\pi(a_{t}|b_{t}, \bm{u}_{t})}\right]
\end{equation}
where $ \rho_{t} = \frac{\pi(a_{t}|b_{t}, \bm{u}_{t})}{\pi_{b}(a_{t}|b_{t})} $ \footnote{$\pi_{b}(a_{t}|b_{t})$ is actually equivalent to $\pi_{b}(a_{t}|s_{t})$, both $b_{t}$ and $s_{t}$ denote the environmental state at an identical moment, depending on whether the agent's notion is represented by belief or fully observable state.} is the importance sampling ratio. Importance sampling mechanisms help ensure that we are not biased by differences between the two policies in choosing actions (e.g. optimal actions may look different than clinical actions taken).

\section{Experimental Results}
We both train and test our algorithm on synthetic and real ICU patients \emph{separately}.
\paragraph{On Synthetic Data}
We first synthesize a dataset where we have full access to its dynamics, from which a true theoretic optimal policy $\pi^*$ can be computed from fully model-reliant approaches such as dynamical programming. Note that although with synthetic data, we choose to learn on an existing (simulated) dataset, to test the algorithm's capability of off-policy learning. The dataset is further divided into two mutually exclusive subsets for algorithm development and test. The suboptimality of our behaviour policy $\pi_b$ that dictates actions during data generation, is systematically related to $\pi^*$ with $\epsilon$-greedy \footnote{$\epsilon$ decides the fraction of occasions when the agent explores actions randomly instead of sticking to the optimal one.}.

In our synthetic data, the action space contains six discrete (categorical) actions. Observed data are generated from a Gaussian mixture model, with state transition probabilities denser at closer states. All parameters for data generation are \emph{unknown} to the reinforcement learning agent.
\begin{figure}[!htp]
  \centering
   \includegraphics[width=\textwidth]{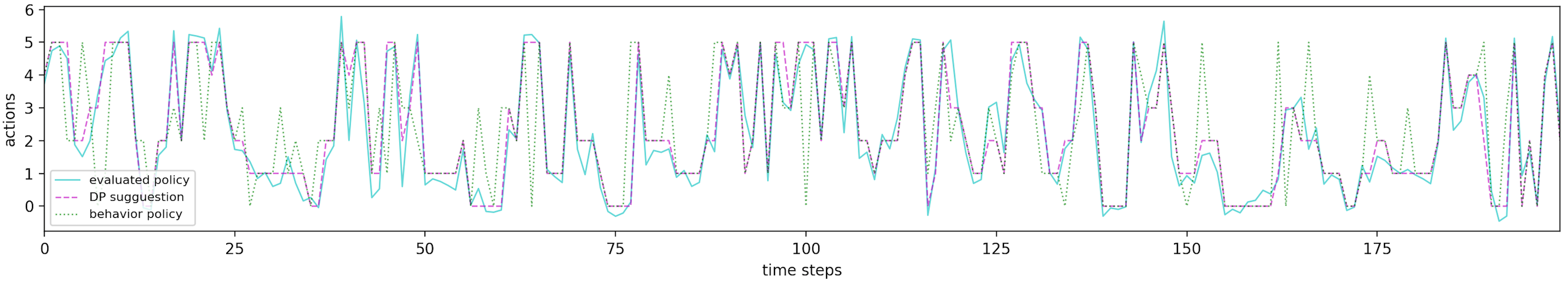}
   \vspace{-0.3cm}
   \caption{Action selections in test set under proposed/optimal/behaviour (showing is $\epsilon=0.3$) policy (synthetic data).}
   \label{fig:act}
\end{figure}

Fig.~\ref{fig:act} visualises action selections under $\pi^*$, $\pi$, and $\pi_b$ of the first 200 time steps (to avoid clutter) in the test set. Note that the optimal and behaviour policies are discrete, while the target policy is continuous. Strict resemblance between the target policy and the optimal policy can be observed. And there is no trace of the target policy varying with the behaviour policy.

\paragraph{On Retrospective ICU Data}
We subsequently apply the methodology on the Medical Information Mart for Intensive Care III (MIMIC-III) \cite{Johnson2017MAF}, a publically available de-identified electronic healthcare record database of patients in ICUs of a US hospital. We include adult patients conforming to the international consensus sepsis-3 criteria \cite{Singer2016}, and exclude admissions where treatment was withdrawn, or mortality was undocumented. This selection procedure leads to 18,919 ICU admissions in total, which are further divided into development and test sets according to proportions 4:1. Time series data are temporally discritised with 4-hour time steps and aligned to the approximate time of onset of sepsis. Measurements within the 4h period are either averaged or summed according to clinical implications. The outcome is mortality, either hospital or 90-day mortality, whichever is available.

The maximum dose of vasopressors (mcg/kg/min) and total volume of intravenous fluids (mL/h) administered 
within each 4h period define our action space. Vasopressors include norepinephrine, epinephrine, 
vasopressin, dopamine and phenylephrine, and are converted to norepinephrine equivalent. Intravenous 
fluids include boluses and background infusions of crystalloids, colloids and blood products, and are normalised 
by tonicity. Patient variables of interest are constituted by demographics (age, gender \footnote{Binary 
variable.\label{bv}}, weight, readmission to ICU \textsuperscript{\ref{bv}}, Elixhauser premorbid 
status), vital signs (modified SOFA, SIRS, Glasgow coma scale, heart rate, systolic/mean/diastolic blood 
pressure, shock index, respiratory rate, Sp$\rm O_2$, temperature), laboratory values (potassium, 
sodium, chloride, glucose, BUN, creatinine, Magnesium, calcium, ionised calcium, carbon dioxide, SGOT, 
SGPT, total bilirubin, albumin, hemoglobin, white blood cells count, platelets count, PTT, PT, INR, pH, Pa$\rm O_2$, PaC$\rm O_2$, base excess, bicarbonate, lactate), ventilation parameters (mechanical 
ventilation \textsuperscript{\ref{bv}}, Fi$\rm O_2$), fluid balance (cumulated intravenous fluid intake, 
mean vasopressor dose over 4h, urine output over 4h, cumulated urine output, cumulated fluid balance 
since admission), and other interventions (renal replacement therapy \textsuperscript{\ref{bv}}, sedation \textsuperscript{\ref{bv}}).

\begin{figure}[!htp]
\centering
		\begin{tabular}{cc}
			\includegraphics[width=0.35\textwidth]{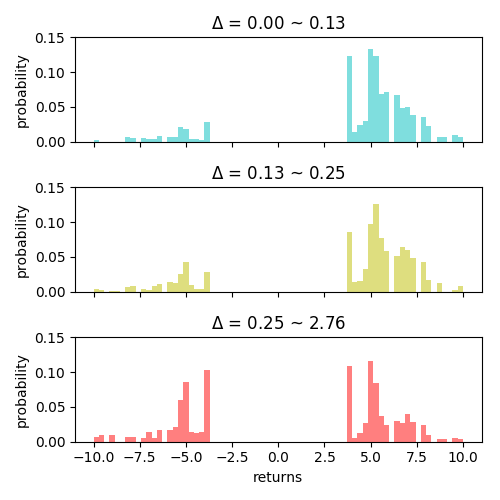}
			&
			\includegraphics[width=0.35\textwidth]{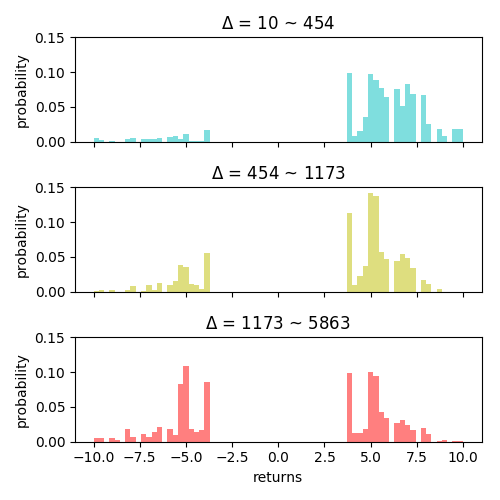}\\
		\end{tabular}
\vspace{-0.3cm}
\caption{
Distributions of returns vs. action deviations.
(Left) distributions of returns for different levels of average absolute vasopressor deviations between clinicians and proposed policy per time step. The uppermost subplot shows empirical outcomes from patients whose vasopressors actually received deviated per time step less than $\frac{1}{3}$ of overall vasopressor deviations (ascending) in the test set, and the lowermost subplot higher than $\frac{2}{3}$; (Right) intravenous fluid counterparts.}
	\label{fig:ret_dev}
\end{figure}
Missing data in patient continuous variables are imputed via linear interpolations, binary variables are interpolated via sample-and-hold. All continuous variables are normalised to $[0, ~1]$.  To promote patient survival (discharge from ICU), each transition to death is penalised by -10, each transition to discharge is rewarded with +10. All non-terminal transitions are zero-rewarded.

Off-policy policy evaluation (OPPE) of the learned policy is usually conferred via importance sampling, where one has to trade between variance and bias. \cite{53jiang2016DROV,54Mahmood2014WISF,55Precup2000ETO,56Thomas2016DataEfficientOP} have extended this to more accurate estimators to minimise estimation error sources for discrete action spaces. However, importance-sampling based approaches usually assume coverage in the behaviour policy $\pi_{b}$ (actions possible in target policy $\pi$ have to be possible in $\pi_{b}$) to calculate the importance sampling ratio, which is mathematically meaningless in our case where both target and behaviour policies are continuous. Instead of using OPE to provide theoretical policy evaluation, we focus on empirically evaluating our learned policy by comparing how the similarity between clinicians' decisions and our suggestions indicates patient outcomes: this provides an empirical validation and is commonly adopted \cite{12Prasad2017MVI,24Nemati2016OptimalMD} for medical scenerios involving retrospective dataset.

\begin{figure}[htp!]
  \centering
   \includegraphics[width=\textwidth]{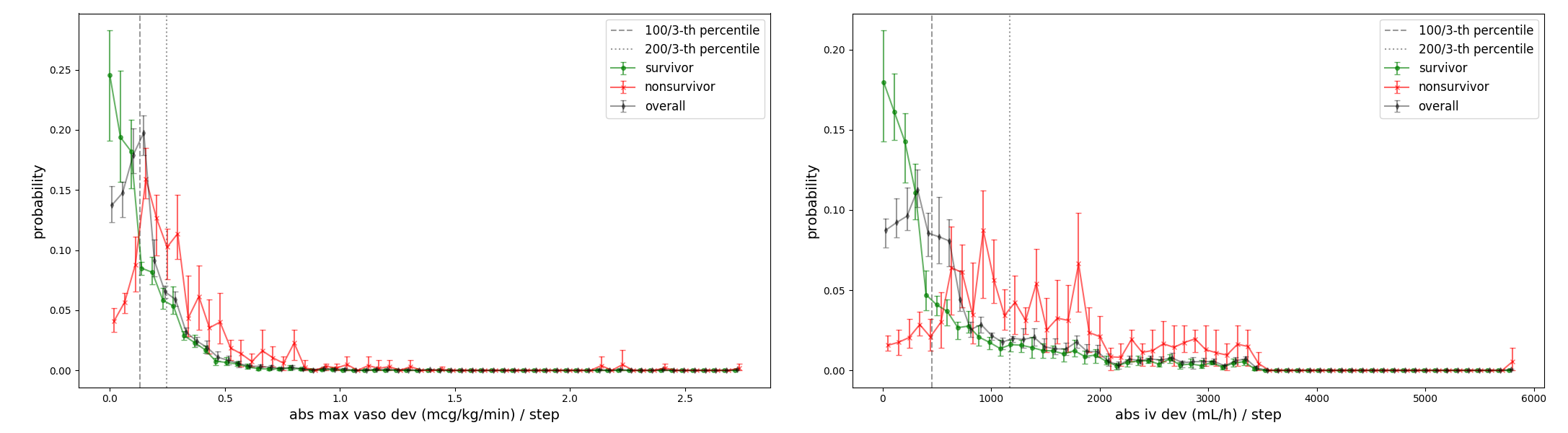}
   \vspace{-0.5cm}
   \caption{Distribution of action deviations with bootstrapping for all survivors, non-survivors, and overall patients in test set are plotted separately, with terciles for each drug plotted as two grey dotted vertical lines, separating the whole test set into three groups with equal patient numbers.}
   \label{fig:mort_boot}
\end{figure}

Fig.~\ref{fig:ret_dev} shows probability mass functions (histograms) of returns of start states (i.e. $ \gamma ^{T-1}r_{T-1} $, $T$ being the length of that time series) in test set divided into three mutually exclusive groups according to the average (per time step) absolute deviation from clinicians' decision and the proposed dose in terms of vasopressor or intravenous fluid within each episode (i.e. individual patient) respectively. The boundaries between two adjacent groups are set to terciles (shown as the grey dotted vertical lines in Fig.~\ref{fig:mort_boot}) of the whole test dataset for each drug to reflect equal weighing. It is observable that for both drugs, higher returns are more likely to be obtained when doctors behave more closely to our suggestions.

The distributions of action deviations between clinicians and the proposed policy in terms of each drug for survivors and non-survivors with bootstrapping (random sampling with replacement) estimations in Fig.~\ref{fig:mort_boot} demonstrate that, among survivors our proposed policy captures doctors' decisions most of the time, while same is not true of non-survivors, especially for intravenous fluids.

\section{Conclusion}
This article provides an online POMDP solution to take into account uncertainty and history information in clinical applications. Our proposed policy is capable of dictating near-optimal dosages in terms of vasopressor and intravenous fluid in a continuous action space, to which behaving similarly would lead to significantly better patient outcomes than that in the original retrospective dataset.

Further research directions include investigating inverse reinforcement learning to recover the reward function that clinicians were conforming to, modelling states/observations to non-trivial distributions to more appropriately extract genuine physiological states, and phrasing the problem into a multi-objective MDP to absorb multiple criteria.

Our overall aim is to develop clinical decision support systems that provision clinicians dynamical treatment planning given previous course of patient measurements and medical interventions, enhancing clinical decision making, not replacing it.




\small


\begin{thebibliography}{10}

\bibitem{gulshan2016development}
V.~Gulshan, L.~Peng, M.~Coram, M.~C. Stumpe, D.~Wu, A.~Narayanaswamy,
  S.~Venugopalan, K.~Widner, T.~Madams, and J.~Cuadros.
\newblock Development and validation of a deep learning algorithm for detection
  of diabetic retinopathy in retinal fundus photographs.
\newblock {\em Jama}, 316(22):2402--2410, 2016.

\bibitem{esteva2017dermatologist}
A.~Esteva, B.~Kuprel, R.~A. Novoa, J.~Ko, S.~M. Swetter, H.~M. Blau, and
  S.~Thrun.
\newblock Dermatologist-level classification of skin cancer with deep neural
  networks.
\newblock {\em Nature}, 542(7639):115, 2017.

\bibitem{litjens2017survey}
G.~Litjens, T.~Kooi, B.~E. Bejnordi, A.~A.~A. Setio, F.~Ciompi, M.~Ghafoorian,
  J.~van~der Laak, B.~van Ginneken, and C.~I S{\'a}nchez.
\newblock A survey on deep learning in medical image analysis.
\newblock {\em Medical image analysis}, 42:60--88, 2017.

\bibitem{10Ernst2006CDB}
D.~Ernst, G.~B. Stan, J.~Goncalves, and L.~Wehenkel.
\newblock Clinical data based optimal sti strategies for hiv: a reinforcement
  learning approach.
\newblock In {\em Proceedings of the 45th IEEE Conference on Decision and
  Control}, pages 667--672, Dec 2006.

\bibitem{11Bothe2013TheUO}
M.~K. Bothe, L.~Dickens, K.~Reichel, A.~Tellmann, B.~Ellger, M.~Westphal, and
  A.~A. Faisal.
\newblock The use of reinforcement learning algorithms to meet the challenges
  of an artificial pancreas.
\newblock {\em Expert review of medical devices}, 10(5):661--73, 2013.

\bibitem{Lowery2013}
C.~Lowery and A.~A. Faisal.
\newblock Towards efficient, personalized anesthesia using continuous
  reinforcement learning for propofol infusion control.
\newblock In {\em 2013 6th International IEEE/EMBS Conference on Neural
  Engineering (NER)}, pages 1414--1417, Nov 2013.

\bibitem{8Shortreed2011ISCD}
S.~M. Shortreed, E.~Laber, D.~J. Lizotte, T.~S. Stroup, J.~Pineau, and S.~A.
  Murphy.
\newblock Informing sequential clinical decision-making through Âreinforcement
  learning: an empirical study.
\newblock {\em Machine Learning}, 84(1):109--136, Jul 2011.

\bibitem{12Prasad2017MVI}
N.~Prasad, L.~Cheng, C.~Chivers, M.~Draugelis, and B.~E Engelhardt.
\newblock A reinforcement learning approach to weaning of mechanical
  ventilation in intensive care units.
\newblock {\em Proceedings of Uncertainty in Artificial Intelligence (UAI)},
  2017.

\bibitem{19Asoh2013AAI}
H.~Asoh, M.~Shiro, S.~Akaho, T.~Kamishima, K.~Hasida, E.~Aramaki, and T.~Kohro.
\newblock An application of inverse reinforcement learning to medical records
  of diabetes treatment.
\newblock In {\em European Conference on Machine Learning and Principles and
  Practice of Knowledge Discovery in Databases}, Sep 2013.

\bibitem{25Lizotte2016MOMD}
D.~J. Lizotte and E.~B. Laber.
\newblock Multi-objective markov decision processes for data-driven decision
  support.
\newblock {\em Journal of Machine Learning Research}, 17(211):1--28, 2016.

\bibitem{24Nemati2016OptimalMD}
S.~Nemati, M.~M. Ghassemi, and G.~D. Clifford.
\newblock Optimal medication dosing from suboptimal clinical examples: A deep
  reinforcement learning approach.
\newblock {\em 2016 38th Annual International Conference of the IEEE
  Engineering in Medicine and Biology Society (EMBC)}, pages 2978--2981, 2016.

\bibitem{29Pineau2003PBVI}
J.~Pineau, G.~J. Gordon, and S.~Thrun.
\newblock Point-based value iteration: An anytime algorithm for pomdps.
\newblock In {\em IJCAI}, pages 1025--1032, 2003.

\bibitem{Bellman1957}
R.~Bellman.
\newblock {\em Dynamic Programming}.
\newblock Princeton University Press, Princeton, NJ, USA, 1 edition, 1957.

\bibitem{62Sutton1999PGM}
R.~S. Sutton, D.~McAllester, S.~Singh, and Y.~Mansour.
\newblock Policy gradient methods for reinforcement learning with function
  approximation.
\newblock In {\em Proceedings of the 12th International Conference on Neural
  Information Processing Systems}, pages 1057--1063, 1999.

\bibitem{30Sondik1978TOCP}
E.~J. Sondik.
\newblock The optimal control of partially observable markov processes over the
  infinite horizon: Discounted costs.
\newblock {\em Operations Research}, 26(2):282--304, 1978.

\bibitem{39Spaan2004APPOMDP}
M.~T.~J. Spaan.
\newblock A point-based pomdp algorithm for robot planning.
\newblock In {\em Proceedings. ICRA '04. 2004 IEEE International Conference on
  Robotics and Automation}, volume~3, pages 2399--2404, April 2004.

\bibitem{33Littman1995LPP}
M.~L. Littman, A.~R. Cassandra, and L.~P. Kaelbling.
\newblock Learning policies for partially observable environments: Scaling up.
\newblock In {\em Proceedings of the Twelfth International Conference on
  International Conference on Machine Learning}, pages 362--370, 1995.

\bibitem{35Cassandra1994AOI}
A.~R. Cassandra, L.~P Kaelbling, and M.~L. Littman.
\newblock Acting optimally in partially observable stochastic domains.
\newblock {\em Twelfth National Conference on Artificial Intelligence
  (AAAI-94)}, pages 1023--1028, 1994.

\bibitem{40Hauskrecht2000VAP}
M.~Hauskrecht.
\newblock Value-function approximations for partially observable markov
  decision processes.
\newblock {\em J. Artif. Int. Res.}, 13(1):33--94, August 2000.

\bibitem{43Paquet06hybridpomdp}
S.~Paquet, B.~Chaib-draa, and S.~Ross.
\newblock Hybrid pomdp algorithms.
\newblock In {\em In Proceedings of The Workshop on Multi-Agent Sequential
  Decision Making in Uncertain Domains (MSDM-2006)}, 2006.

\bibitem{44McAllester1999APF}
D.~A. McAllester and S.~Singh.
\newblock Approximate planning for factored pomdps using belief state
  simplification.
\newblock In {\em Proceedings of the Fifteenth Conference on Uncertainty in
  Artificial Intelligence}, pages 409--416, 1999.

\bibitem{45Kearns2002SSA}
M.~Kearns, Y.~Mansour, and A.~Y. Ng.
\newblock A sparse sampling algorithm for near-optimal planning in large markov
  decision processes.
\newblock {\em Mach. Learn.}, 49(2-3):193--208, November 2002.

\bibitem{42Smith2004HSV}
T.~Smith and R.~Simmons.
\newblock Heuristic search value iteration for pomdps.
\newblock In {\em Proceedings of the 20th Conference on Uncertainty in
  Artificial Intelligence}, pages 520--527, 2004.

\bibitem{51Washington97bi-pomdp:bounded}
R.~Washington.
\newblock Bi-pomdp: Bounded, incremental partially-observable markov-model
  planning.
\newblock In {\em In Proceedings of the 4th European Conference on Planning
  (ECP}, pages 440--451. Springer, 1997.

\bibitem{52Ross2007AAO}
S.~Ross and B.~Chaib-Draa.
\newblock Aems: An anytime online search algorithm for approximate policy
  refinement in large pomdps.
\newblock In {\em Proceedings of the 20th International Joint Conference on
  Artifical Intelligence}, pages 2592--2598, 2007.

\bibitem{Schwarz1978BIC}
E.~S. Gideon.
\newblock Estimating the dimension of a model.
\newblock {\em The Annals of Statistics}, 6, 03 1978.

\bibitem{Murphy2012MLA}
K.~P. Murphy.
\newblock {\em Machine learning: a probabilistic perspective}.
\newblock Cambridge, MA, 2012.

\bibitem{63buntine2012ABVOP}
W.~Buntine and M.~Hutter.
\newblock A bayesian view of the poisson-dirichlet process.
\newblock {\em arXiv:1007.0296v2}, 2012.

\bibitem{Klopf1986}
A.~H. Klopf.
\newblock A drive‐reinforcement model of single neuron function: An
  alternative to the hebbian neuronal model.
\newblock {\em AIP Conference Proceedings}, 151(1):265--270, 1986.

\bibitem{9Sutton98a}
R.~S. Sutton and A.~G. Barto.
\newblock {\em Reinforcement Learning : An Introduction}.
\newblock MIT Press, 1998.

\bibitem{66Degris2012OffPolicyA}
T.~Degris, M.~White, and R.~S. Sutton.
\newblock Off-policy actor-critic.
\newblock {\em ICML}, 2012.

\bibitem{Johnson2017MAF}
A.~E.~W. Johnson, T.~J. Pollard, L.~Shen, L.~H. Lehman, M.~Feng, M.~Ghassemi,
  B.~Moody, P.~Szolovits, L.~Anthony~Celi, and R.~G. Mark.
\newblock Mimic-iii, a freely accessible critical care database.
\newblock 3, 2017.

\bibitem{Singer2016}
M.~Singer, C.S. Deutschman, C.~Seymour, et~al.
\newblock The third international consensus definitions for sepsis and septic
  shock (sepsis-3).
\newblock {\em JAMA}, 315(8):801--810, 2016.

\bibitem{53jiang2016DROV}
N.~Jiang and L.~Li.
\newblock Doubly robust off-policy value evaluation for reinforcement learning.
\newblock In {\em Proceedings of The 33rd International Conference on Machine
  Learning}, volume~48, pages 652--661, Jun 2016.

\bibitem{54Mahmood2014WISF}
A.~R. Mahmood, H.~P. van Hasselt, and R.~S. Sutton.
\newblock Weighted importance sampling for off-policy learning with linear
  function approximation.
\newblock In {\em Advances in Neural Information Processing Systems 27}, pages
  3014--3022. 2014.

\bibitem{55Precup2000ETO}
D.~Precup, R.~S. Sutton, and S.~P. Singh.
\newblock Eligibility traces for off-policy policy evaluation.
\newblock In {\em Proceedings of the Seventeenth International Conference on
  Machine Learning}, pages 759--766, 2000.

\bibitem{56Thomas2016DataEfficientOP}
Philip~S. Thomas and Emma Brunskill.
\newblock Data-efficient off-policy policy evaluation for reinforcement
  learning.
\newblock In {\em ICML}, 2016.

\end{thebibliography}
\end{document}